\documentclass[authorversion]{sigchi}
\usepackage{algorithm}
\usepackage{algorithmic}
\usepackage{amsmath}

\newcommand{\argmax}{\mathop{\rm arg~max}\limits}
\newcommand{\addition}[1]{#1}

\usepackage{balance}       
\usepackage{graphics}      
\usepackage[T1]{fontenc}   
\usepackage{txfonts}
\usepackage{mathptmx}
\usepackage[pdflang={en-US},pdftex]{hyperref}
\usepackage{color}
\usepackage{booktabs}
\usepackage{textcomp}

\usepackage{microtype}        
\usepackage{ccicons}          

\usepackage{todonotes}

\def\plaintitle{Autonomous Self-Explanation of Behavior \\for Interactive Reinforcement Learning Agents}

\def\emptyauthor{}
\def\plainkeywords{Human Robot Cooperation; Interactive Reinforcement Learning; Instruction-based Behavior Explanation}

\makeatletter
\def\url@leostyle{%
  \@ifundefined{selectfont}{
    \def\UrlFont{\sf}
  }{
    \def\UrlFont{\small\bf\ttfamily}
  }}
\makeatother
\def\pprw{8.5in}
\def\pprh{11in}

\definecolor{linkColor}{RGB}{6,125,233}
\hypersetup{%
  pdftitle={\plaintitle},
  pdfauthor={\emptyauthor},
  pdfkeywords={\plainkeywords},
  pdfdisplaydoctitle=true, 
  bookmarksnumbered,
  pdfstartview={FitH},
  colorlinks,
  citecolor=black,
  filecolor=black,
  linkcolor=black,
  urlcolor=linkColor,
  breaklinks=true,
  hypertexnames=false
}

\begin{document}

\title{\plaintitle}

\numberofauthors{2}
\author{
	\alignauthor{Yosuke Fukuchi$^1$\\
    \email{fukuchi@ailab.ics.keio.ac.jp}}\\
	\alignauthor{Masahiko Osawa$^{1 2}$\\
	\email{mosawa@ailab.ics.keio.ac.jp}}\\
\and
	\alignauthor{\vspace{-1.5em}Hiroshi Yamakawa$^{3 4}$\\
	\email{hiroshi\_yamakawa@dwango.co.jp}}\\
	\alignauthor{\vspace{-1.5em}Michita Imai$^1$\\
    \email{michita@ailab.ics.keio.ac.jp}}\\
}

\CopyrightYear{2017} 
\setcopyright{authorversion} 
\conferenceinfo{HAI '17,}{October 17--20, 2017, Bielefeld, Germany}
\isbn{978-1-4503-5113-3/17/10}
\acmPrice{\$15.00}
\doi{https://doi.org/10.1145/3125739.3125746}

\maketitle

\begin{abstract}
	In cooperation, the workers must know how co-workers behave. 
	However, an agent's policy, which is embedded in a statistical machine learning model, is hard to understand,
	and requires much time and knowledge to comprehend.
	Therefore, it is difficult for people to predict the behavior of machine learning robots, 
	which makes Human Robot Cooperation challenging.
	In this paper, we propose Instruction-based Behavior Explanation (IBE), a method to explain an autonomous agent's future behavior.
	In IBE, an agent can autonomously acquire the expressions to explain its own behavior by reusing
	the instructions given by a human expert to accelerate the learning of the agent's policy.
	IBE also enables a developmental agent, whose policy may change during the cooperation, 
	to explain its own behavior with sufficient time granularity.
\end{abstract}


\footnotetext[1]{Keio University,  Yokohama, Japan}
\footnotetext[2]{Research Fellow of Japan Society for the Promotion of Science, Tokyo, Japan}
\footnotetext[3]{Dwango Artificial Intelligence Laboratory, Tokyo, Japan}
\footnotetext[4]{The Whole Brain Architecture Initiative, Tokyo, Japan}

\section{Introduction}
Human Robot Cooperation (HRC), in which people and robots work on the same task together in a shared environment, is an effective concept for both industrial and domestic robots \cite{amor2014interaction, dimeas2016online}.
By working in a complementary manner, both robots and people overcome the disadvantages of each other in order to achieve difficult tasks that cannot be achieved by either of them. 
Cooperative robots require the ability to deal with complicated real-world information, 
and machine learning technics, such as deep reinforcement learning (DRL), are expected to realize the real-world information processing.

In cooperation, the workers must know how the other co-workers behave, in order to avoid dangerous misunderstandings and decide what roles to take in the situation \cite{hayes2013challenges}.
However, it is difficult for people to predict the machine learning agent's behavior.
The control logic embedded in a statistical machine learning model, especially in a deep learning model, 
is incomprehensible for most people, and requires much time and knowledge to understand.

Previous studies have shown that the interaction between people and robots develops the people's understandings of robots and improves the performance of the cooperation.
Andresen et al. proposed a robot that projects the robot's own intentions and instructions on real-world objects \cite{7745145}.
This robot detects the locations and shapes of nearby objects, and projects information directly on them.
The projection improved the effectiveness and efficiency of collaboration tasks.
Hayes et al. proposed an answering system that explains autonomous agents' policy \cite{hayes2017improving}.
This system builds a statistical model for the autonomous agent's actions,
and deals with some sets of natural language questions based on the model, so that people can get insights into the control logic of the agent. 
The proposed method successfully summarized the policies of both hand-coded and machine learning agents in some domains, regardless of internal representation of the agent's control logic.

\begin{figure}[tbp]
  \begin{center}
  	\includegraphics[clip,width=\linewidth]{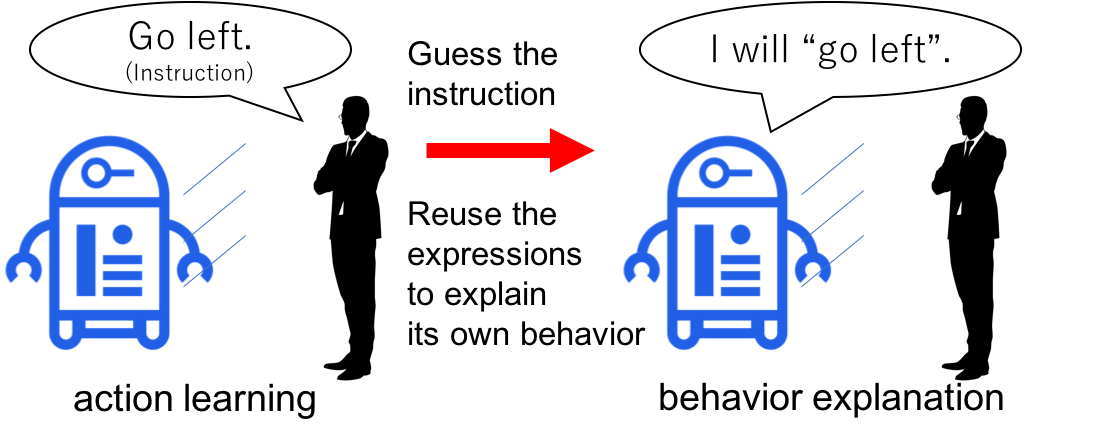}
  	\caption{Instruction-based Behavior Explanation}
  	\label{fig:1}
  \end{center}
\end{figure}

However, all the information provided by the projection robot is designed by programmers.
More complex behaviors acquired autonomously by machine learning technics make it impossible for people to
design information to show their co-workers.
The natural language answering system also requires designers to prepare a mapping from the agent's action to a communicable predicate that explains the action.
In addition, Hayes et al. assumed that the policy of an agent does not change after building a model of the agent's behavior;
therefore, this method does not work on a developmental agent that renews its policy gradually in actual human robot collaboration.
Moreover, the work attempts to assign a communicable predicate to an action in one time step;
however, explaining one step action is usually fine-spun when we consider a machine learning model that controls complex behavior of a robot.
An agent's behavior that people can recognize is the result of a sequence of actions.
In order to explain the behavior of an agent to people,
we have to consider the agent's actions with longer time granularity.

This paper proposes Instruction-based Behavior Explanation (IBE), which is a method that explains the future behavior of a reinforcement learning agent in any situation.
In IBE, we consider a setting of Interactive Reinforcement Learning (IRL) \cite{thomaz2005real}.
IRL is a framework in which a machine learning agent receives expert's instructions to accelerate the agent's policy acquisition. 
The IBE reuses the instruction as representations to explain an agent's behavior (Fig. \ref{fig:1}).
However, in contrast to IRL, the designer \addition{or the instructor} of agents does not have to give the relationship between instructions and agent's actions \addition{explicitly}.
In IBE, an agent guesses the meanings of instructions on the assumption that, when an agent receives more rewards, it is more likely that the behavior of the agent followed the instruction.
With this assumption, an agent can autonomously acquire the expression to explain the behavior. 
Besides, IBE estimates an agent's behavior by simulating the transitions of the environment in each time step.
The successive simulations make it possible to deal with a developmental agent whose policy is changeable.
Moreover, by broadening the time span of the simulation, IBE can output information with sufficient time granularity.

\begin{figure}[tbp]
  \begin{center}
    \includegraphics[clip,width=\linewidth]{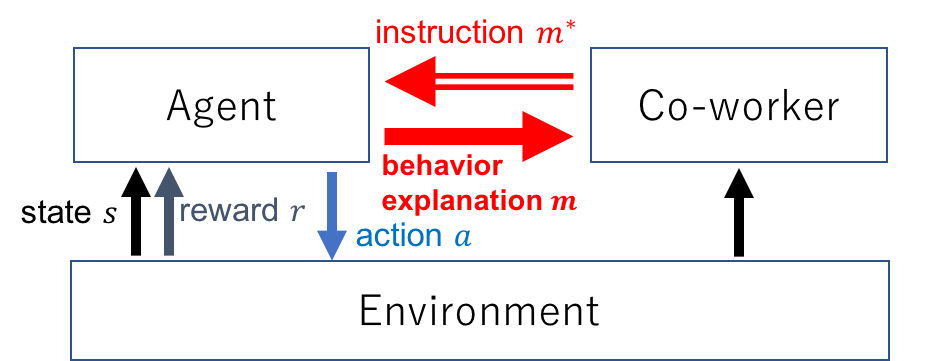}
	  \caption{The settings of IBE}
    \label{fig:whole}
  \end{center}
\end{figure}

\section{background}
\subsection{Reinforcement Learning (RL)}
RL is a type of learning, which acquires an agent's policy autonomously in a sequential decision making process \cite{sutton1998reinforcement}.
An agent observes the state of the environment $s_t$ and selects an action $a_t$ in time $t$.
The state of the environment changes to $s_{t+1}$ by the agent's action $a_t$, and the agent receives a reward $r_t$ from the environment.
An agent decides the action based on its policy $\pi$, where $\pi(s, a)$ is the probability of the agent to take an action $a$ in the environment state $s$.
The goal of an RL agent is to find the optimal policy $\pi^*$ that maximizes the total reward $R$.
In this paper, we consider an agent that learns its policy with RL.

\subsection{Interactive Reinforcement Learning (IRL)}
In a complex situation in which the state spaces and action spaces are very large, 
the learning process of an RL becomes excessively long \cite{knox2009interactively}.
In order for a cooperative agent to acquire its policy in the real-world with machine learning technics,
it is necessary to deal with an increase in the search time.
IRL is an approach that can solve the search time problem.
In IRL, a human or an agent expert instructs a beginner agent in real time so that the beginner can learn the policy efficiently \cite{cruz2016training}.
Narrowing down the search spaces with the instruction can also help a cooperative agent learn the policy in the real-world.

Therefore, in this study, we consider a scenario in which a human expert instructs to an agent (Fig. \ref{fig:whole}).
An expert gives an instruction signal $m^*_t$ to a beginner agent for every time step.
$m^*_t$ is a real number, which represents an instruction from an expert to an agent.
\addition{
The instruction spaces are much narrower than the state spaces. Therefore, it is expected that the instructions can be linked to the actions
more quickly than the environment states while agents will be able to select actions 
with the environment state information more delicately because the instruction signal is less informative than the state information.
}

\section{Instruction-based Behavior Explanation (IBE)}
\begin{figure}[tbp]
  \begin{center}
    \includegraphics[clip,width=0.9\linewidth]{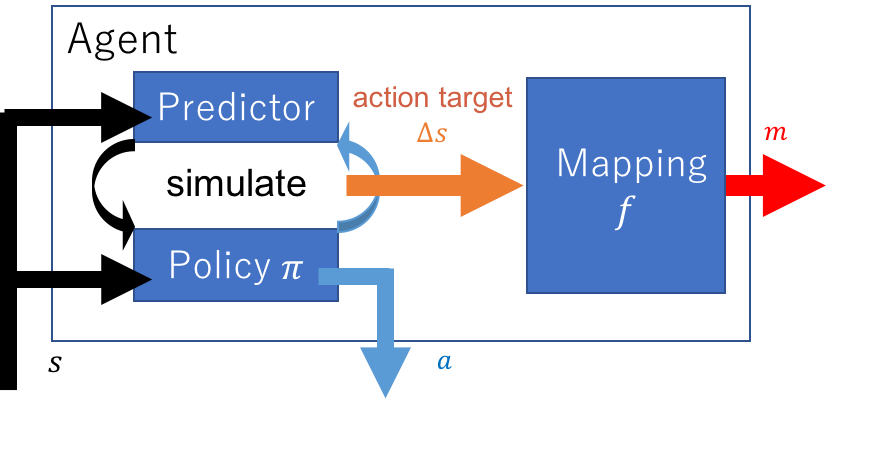}
	  \caption{The flow to explain an agent's behavior in IBE}
    \label{fig:detail}
  \end{center}
	\vspace{-1.6em}
\end{figure}

In this paper, we propose IBE, a method to explain an autonomous agent's behavior with the expressions given by a human expert as instruction (Fig. \ref{fig:detail}).
IBE consists of two steps: (i) estimating the target of the agent's actions by simulation and (ii) acquiring a mapping from the target of the agent's actions to the expressions, in order to explain the action target based on the instruction signal given by a human expert.

\subsection{Estimation of the action target}
In this study, we define the target of the agent's actions at time $t$ as a change in the environment state after the agent's actions in $n$ steps $\Delta s_t$.
\begin{equation}
	\Delta s_t = s_{t+n} - s_t
\end{equation}
With the introduction of the time span $n$, IBE can output the behavior explanation with human-understandable time granularity. 

IBE estimates $\Delta s_t$ with the agent's policy $\pi (s,a)$ and $Predictor(s, a)$ (Algorithm \ref{alg1}).
\addition{
$Predictor$ predicts the environment state in the next time step $s_{t+1}$ with current state $s_t$ and agent's action $a_t$ as input.
}
\begin{algorithm}[tbp]
\caption{Estimation of the action target}
\label{alg1}
\begin{algorithmic}
	\REQUIRE 
		$s_t$: current state, 
		$n$: range of the time steps 
	\ENSURE $\Delta s_t$: transition of the environment state
	
	\STATE $s = s_t$;
	\FOR{ $counter = 1$ to $n$}
		\STATE $a = \argmax_a \pi (s,a)$;
		\STATE $s = Predictor(s,a)$;
    \ENDFOR
	\STATE3$\Delta s_t = s - s_t$;
	\STATE return $\Delta s_t$;

\end{algorithmic}
\end{algorithm}

\subsection{Mapping from the agent's behavior to the explanation signal}
Next, the IBE decides an expression $m_t$ to explain the change in the environment $\Delta s_t$.
In other words, we consider the mapping $f: \Delta s \rightarrow m$.

\addition{
By autonomously acquiring the relationship between instructions and states of the environment,
we can also use the relationship to accelerate the learning process.
For example, we will be able to judge whether the agent followed the instructions and give additional feedback to the agent's actions.
}

First of all, we collect the history of the environment state $s_t$, rewards $r_t$, and the instruction at the time $m^*_t$.
Then we choose the episodes in which the total reward is top $x$, and calculate the changes in the environment caused by the agent's actions $\Delta s_t$.
After that, we divide the sets of $\Delta s$ into clusters $C^1, C^2, ..., C^k$ with a clustering method.
The classifier acquired in the clustering process can divide any $\Delta s$ into a cluster.
The mapping $f$ is obtained by determining the explanation for each cluster ($m^1, m^2, ..., m^k$).
\begin{equation}
	f(\Delta s_t) = m^i  (\Delta s_t \in C^i)
\end{equation}
$M^{*i}$ is a set of instructions accompanied by $\Delta s \in C^i$.
\begin{equation}
	M^{*i} = \{ m^*_t | s_t \in C^i \}
\end{equation}
To decide the explanation $m$, we assume that it is more likely for the agent to have followed an expert's instruction when the agent received more rewards.
We calculate the expected values of the instructions $M^{*i}$ for each cluster.
\begin{equation}
	e^i = E[m^* | m^* \in M^{*i}] 
\end{equation}
Finally, we normalize $e^i$ between the clusters to be $m^i$.

\section{Case study}
\begin{figure}[tbp]
  \begin{center}
    \includegraphics[clip,width=0.49\linewidth]{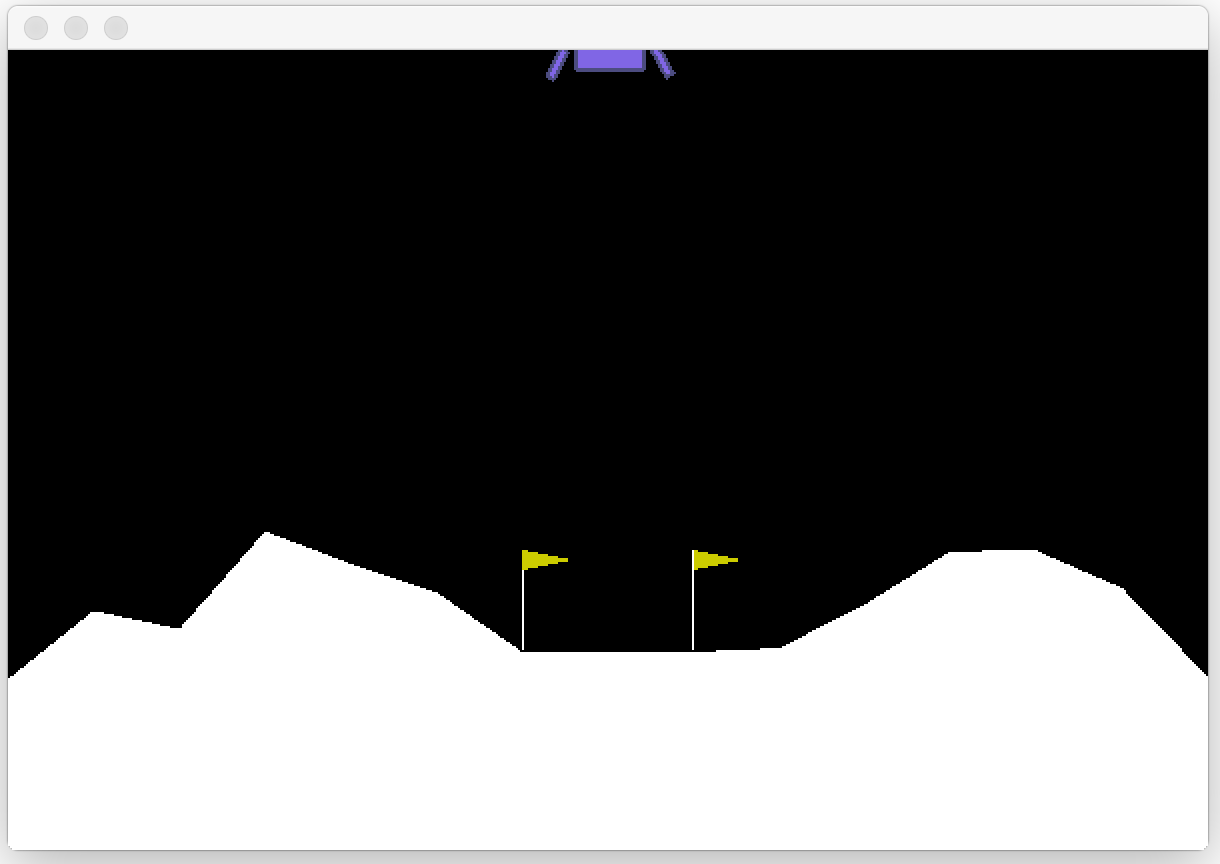}
    \includegraphics[clip,width=0.49\linewidth]{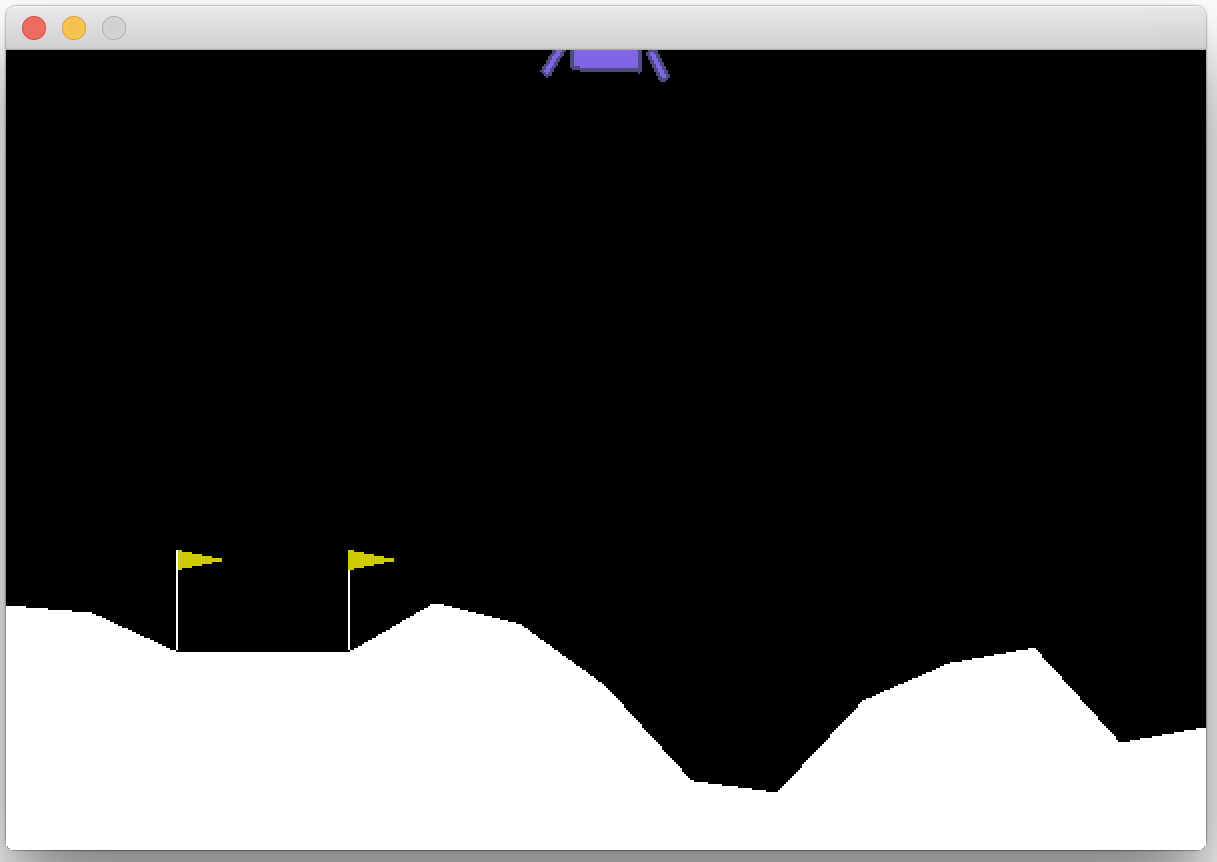}
    \caption{
		Modified Lunar Lander v2. An agent receives the location of the goal with the environment state, and 
		learns to soft-land on the goal.
	}
    \label{fig:game}
  \end{center}
\end{figure}

We constructed a game environment based on Lunar Lander v2, which was released on Open AI gym \cite{brockman2016openai}
to evaluate the IBE.
The goal of the game is to soft-land a rocket on the moon (Fig. \ref{fig:game}).
The available actions $a$ are as follows: do nothing, fire left orientation engine, fire main engine, and fire right orientation engine.
The landing pad of the original Lunar Lander v2 is always in the center; however, we randomly changed the landing location to the left, center, and right, to make it more difficult for people to anticipate the agent's behavior. 
In every time step, the reward for a rocket agent is calculated based on five parameters: the distance to the goal, the speed of the rocket, the degree of inclination,
the amount of use of fuel, and whether the legs of rockets are grounded on the moon.

We prepared two rocket agents with deep reinforcement learning models \cite{mnih2015human}.
Agent A is in the middle of the policy acquisition, and possibility to soft-land on the goal is 63.3\%.
Agent B has better policy than Agent A, and possibility is 83.3 \%.
\addition{
We prepared agent A in order to consider the applicability of the IBE to agents in an earlier stage of action acquisition.
Unskilled agents can act unexpected immature behavior, so it is better to be able to explain the agent's behavior as soon as possible.
}

We decided the instruction $m_t$ for an agent as given by formula \ref{eqn:instruction}.
\begin{equation}
	\label{eqn:instruction}
	m^*_t = \left \{
		\begin{array}{ll}
			-1 & \text{if the agent is flying right of the right flag} \\
			+1 & \text{else if the agent is flying left of the left flag} \\
			0 & \text{otherwise}
		\end{array}
	\right.
\end{equation}
$m = -1, 0, 1$ means "Fall to the left," "Fall straight down," and "Fall to the right," respectively.

\addition{
In the experiments, we prepared the predictor module using the same game engine as the Lunar Lander v2 to eliminate the uncertainty of the prediction.
}

\subsection{Preliminary Experiment}
\begin{figure}[tbp]
  \begin{center}
    \includegraphics[clip,width=0.49\linewidth]{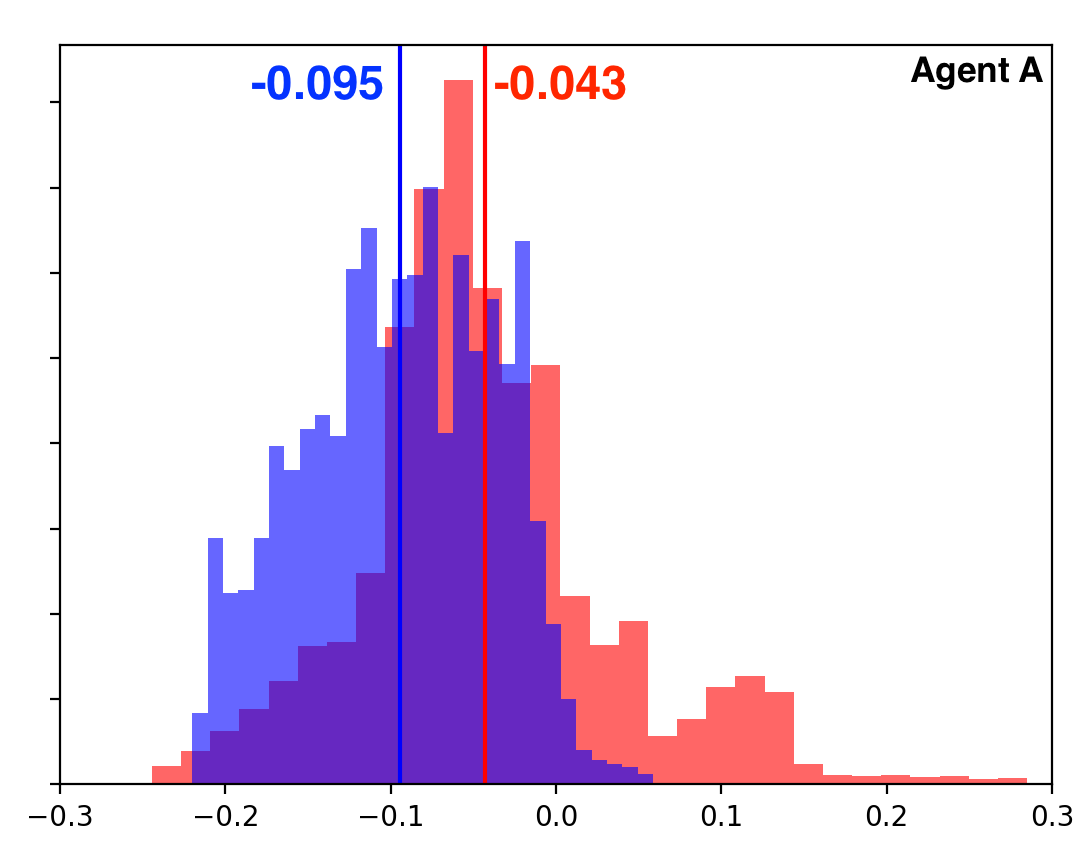}
    \includegraphics[clip,width=0.49\linewidth]{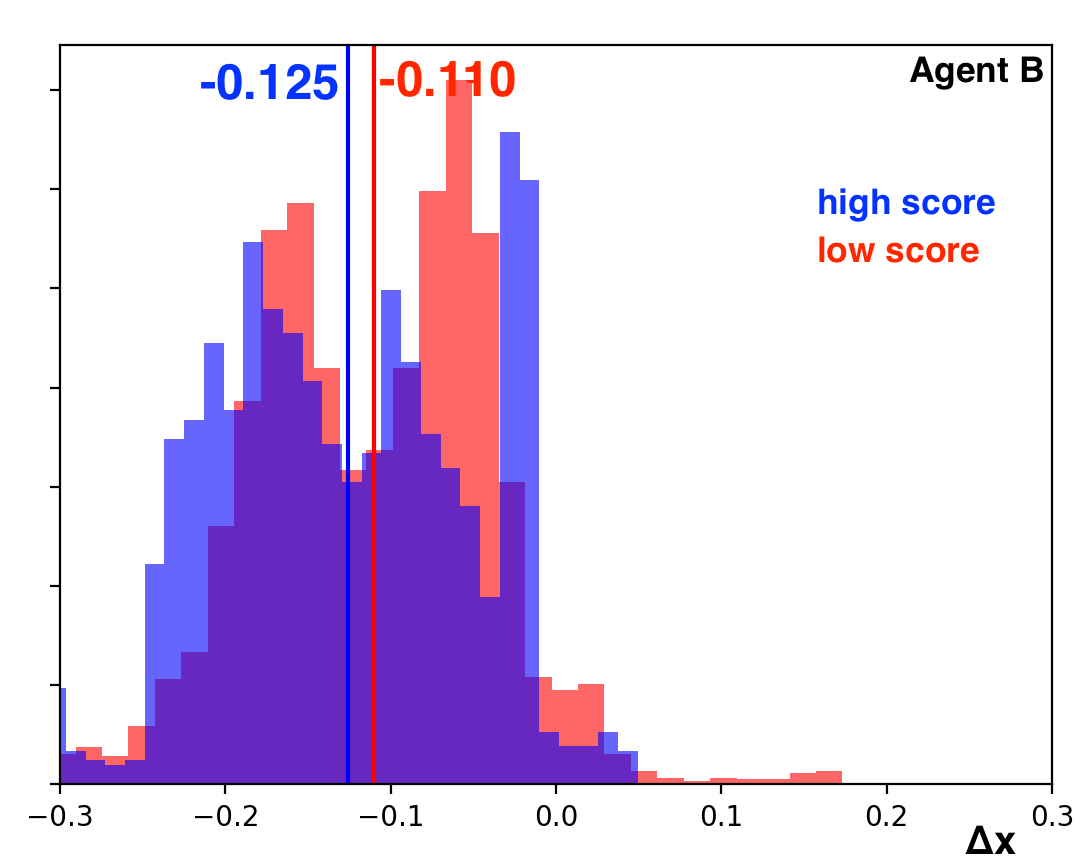}
	  (a) $m_t = -1$ ("Fall to the left") \\
    \includegraphics[clip,width=0.49\linewidth]{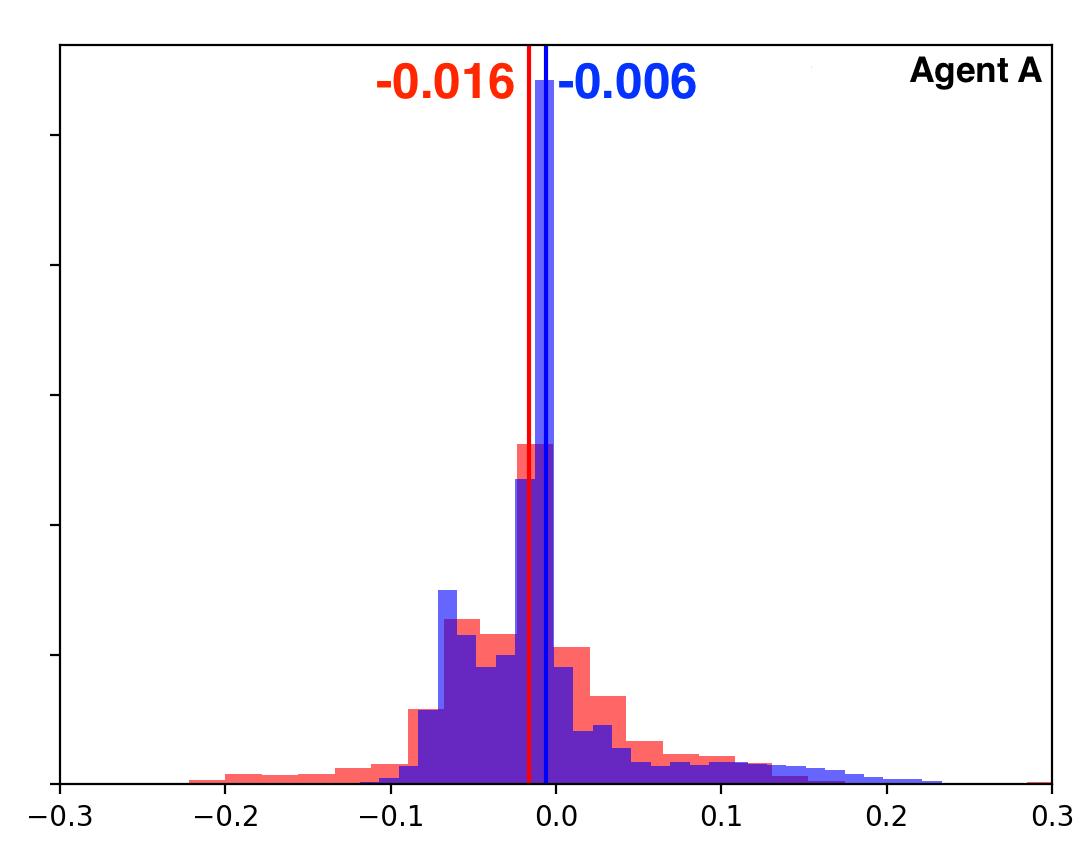}
    \includegraphics[clip,width=0.49\linewidth]{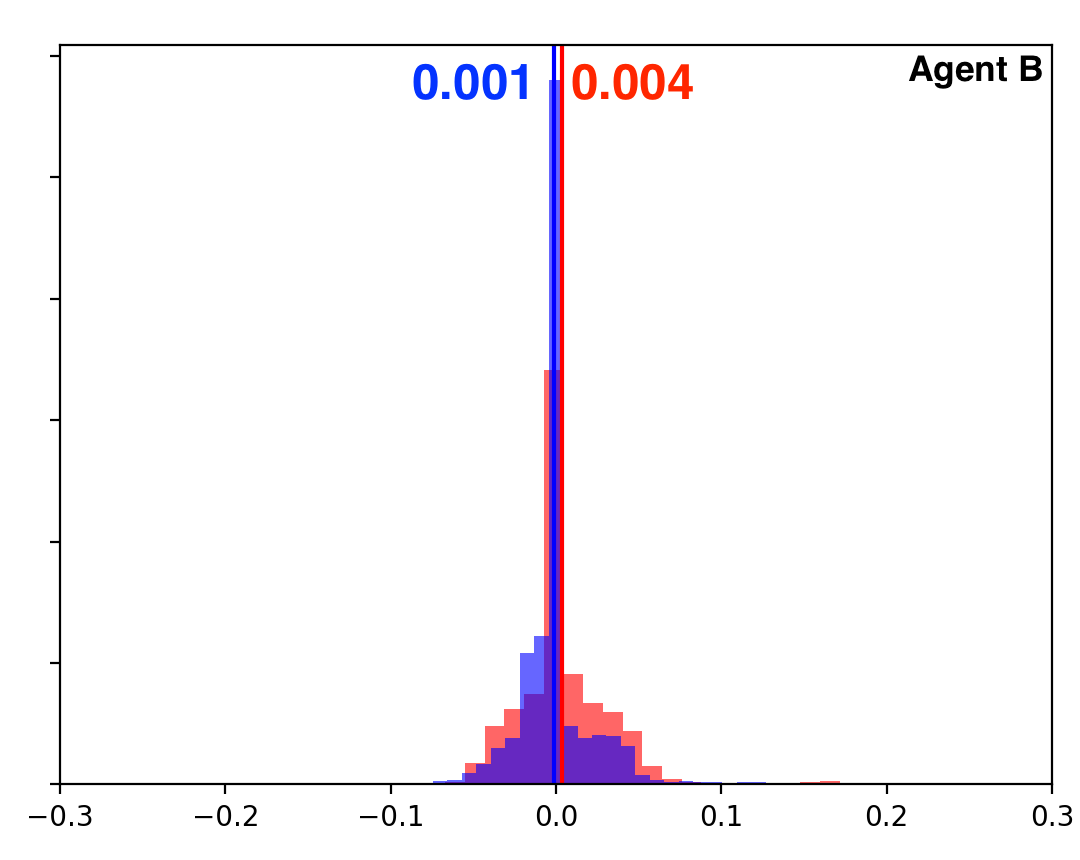}
		(b) $m_t = 0 $ ("Fall straight down") \\
	  \caption{
		The histogram of $\Delta x$ in which the agent received the instructions.
		$\Delta x$ is negative when the agent moves left, and zero when the agent falls straight down.
		The vertical bars indicate the average values of $\Delta x$.
		In (a), the blue areas, which indicates the high-score episodes, are distributed to the left side more than the red areas, which indicates the low-score episodes.
		The result shows that, with the assumption, the IBE can extract the agent's behavior which follows the expert's instructions.
	}
    \label{fig:hist}
  \end{center}
\end{figure}

Firstly, we inspected the assumption that when an agent received more rewards, it is more likely that the behavior of the agent followed the instruction in the Lunar Lander.
The histogram in Fig. \ref{fig:hist} shows the probability distribution of the amount of the agents' moves in the horizontal direction $\Delta x = x_{t+n} - x_t$, when the experts told the agents to "fall to the left" ($m_t=-1$) and "fall straight down" ($m_t=0$).
$\Delta x$ is negative when the agent moves left, and zero when the agent falls straight down.
We divided the episodes into two groups: episodes whose total reward is the top 25 \%, and the others.
The time spun for the simulation $n = 60$.

Fig \ref{fig:hist} shows 
that in the high-score episodes, agents follow the instruction more often than in the low-score episodes, especially for the low-score agent (Agent A).
The result shows that considering the total reward in an episode helps in extracting the the agents' behavior that follows the expert's instructions. 

\subsection{Prediction Task}
We conducted an experiment to inspect the effect of explanation of an agent's behavior by IBE.
We flattened the ground and got rid of the flags so that the participants could not know where the goal is, and recorded the game scenes of agent B.
Then we selected 20 episodes whose length was more than 80 frames, and cut out 80 frames until landing.
The participants of the experiment watched the first 20 frames to predict where the agent landed with or without the explanation by IBE, 
and then checked the actual behavior of the agent.
\addition{
Nine male students aged 21 to 28 who had never played the game participated in the experiment. 
We showed the output of the IBE to five of the participants, and the others predicted the landing spot without the output. 
}

\begin{figure}[t!]
  \begin{center}
    \includegraphics[clip,width=0.6\linewidth]{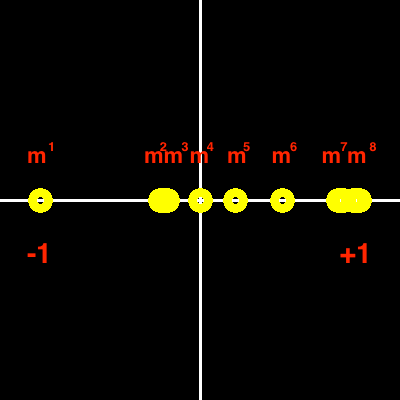}
    \caption{$m^i$ to explain the target of the action of an agent in a cluster $C^i$.
		The distribution of $m^i$ is biased to the right.
		IBE can explain the movements to the right in more detail than left because of the bias.
	  }
    \label{fig:cluster}
  \end{center}
\end{figure}

The number of clusters $k$ was eight, and the time spun for the simulation $n$ was 60.
The outputs of IBE was normalized between -1 to 1, and visualized as shown in Fig. \ref{fig:cluster}.
The $Predictor(s, a)$ of IBE was Box2d \cite{catto2011box2d}, which is the same 2D game engine as Lunar Lander v2.
Before the clustering of the change in the environment $\Delta s$, we normalized $\Delta s$ and used k-means clustering.

\subsection{Result}
\begin{figure}[t!]
  \begin{center}
    \includegraphics[clip,width=0.75\linewidth]{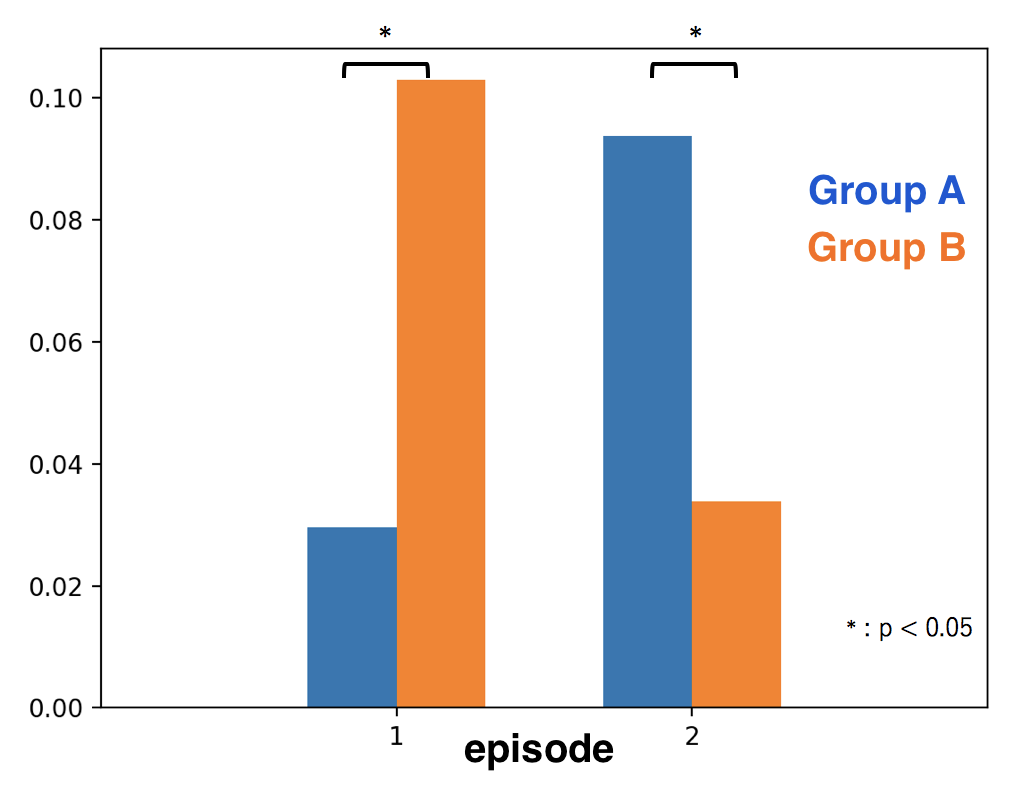}
    \caption{
		The percentage of error against the width of the game in episode 1 and 2.
	  }  
	\label{fig:bar}
  \end{center}
\end{figure}

\begin{figure}[t!]
  \begin{center}
    \includegraphics[clip,width=0.9\linewidth]{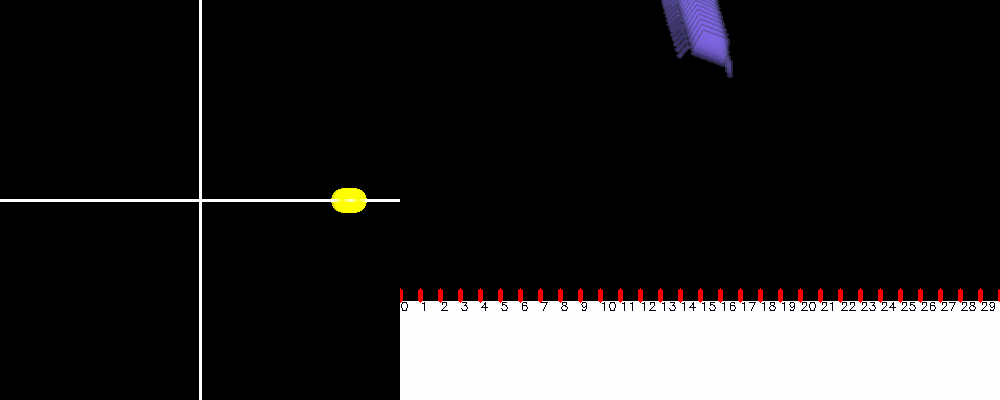}
	\includegraphics[clip,width=0.9\linewidth]{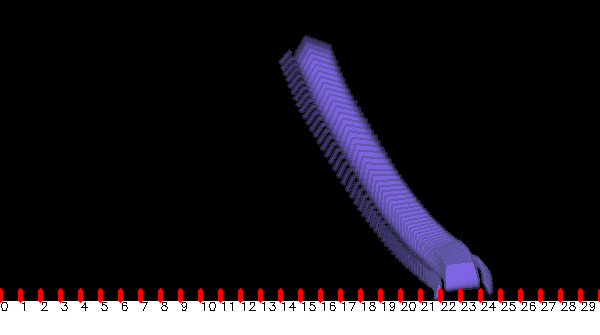}
    \caption{Explanation was effective in episode 1.
	 	The upper right portion shows agent B's behavior in the first 20 frames.
		The upper left portion shows the visualization of the output of IBE in the first 20 frames.
		The lower portion shows the agent B's behavior in the last 60 frames.
		At first agent B fell linearly, but then bent the course to the right widely.
	 }
    \label{fig:result1}
  \end{center}
\end{figure}

\begin{figure}[t!]
  \begin{center}
    \includegraphics[clip,width=0.9\linewidth]{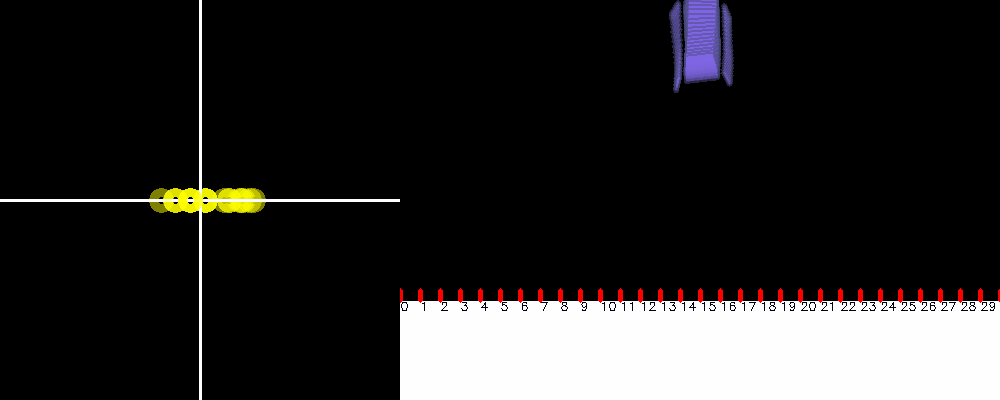}
    \includegraphics[clip,width=0.9\linewidth]{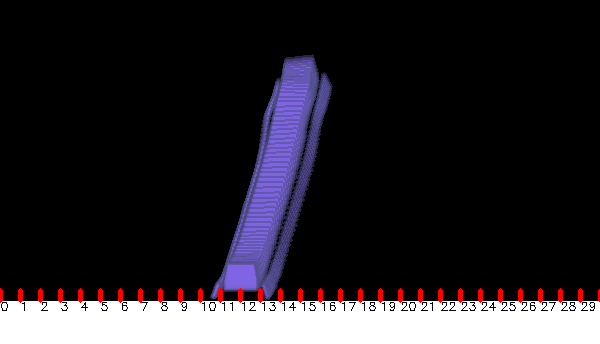}
    \caption{Explanation misled the participants in episode 2.
	  Agent B went a little to the left, but the outputs of IBE gathered in the center,
	  which misled the participants to think the rocket fell straight down.
	  }
    \label{fig:result2}
  \end{center}
	\vspace{-1em}
\end{figure}

We compared the accuracy of the participants' predictions.
The calculation of T-test confirmed significant differences between the participants with the IBE's explanation (group A) and without the explanation (group B) 
in two episodes (Fig. \ref{fig:bar}).
In episode 1 (Fig. \ref{fig:result1}) group A was significantly more accurate than group B,
and episode 2 was the reverse of episode 1 (Fig. \ref{fig:result2}).

The IBE generated complementary expressions from the three expressions given by the experts (Fig. \ref{fig:cluster}).
The complementary expression makes it possible to explain the degree of agent's behavior, whereas the expert's instruction does not.
In the first 20 frames of episode 1, the rocket agent fell linearly, but deviated greatly to the right.
We can say that in episode 1, it is difficult for group B to predict the rocket's behavior, 
because the movements of the agent in the first 20 frames and the last 60 frames are quite different. 
However, the outputs of IBE was stuck to the right; therefore, the participants in group A could anticipate that the agent moved to the right considerably.

On the other hand, Fig. \ref{fig:cluster} shows that the clusters are unevenly distributed.
Since the clusters are skewed to the right, it is possible that the resolution of the explanation was low when the agent moved to the left.
In other words, IBE can explain the movements to the right in more detail than left.
In the episode 2, rocket gently fell to the left (Fig. \ref{fig:result2}); however, 
IBE did not output $m^1$. The output of IBE gathered around zero.
Therefore, the participants misunderstood that the rocket fell straight down.
The result suggests that we need to consider how to divide the environment change $\Delta s$ to assign an explanation signal.

\section{Conclusion}
This paper proposed Instruction-based Behavior Explanation, a method to guess the meaning of an expert's instruction and reuse the expression of the instruction to explain the agent's behavior.	
With IBE, the designer of an agent does not have to prepare a mapping from the agent's behavior to an expression, in order to explain the behavior.
By simulating the agent's behavior, we can deal with a developmental agent whose policy changes during the interaction with the environment. 
Simulating the long-spun behavior of an agent also makes it possible to explain an agent's behavior with sufficient time granularity.
The results of the experiments showed the partial contribution that the explanation autonomously acquired by the IBE enriched people's understandings of the agent's future behavior.

\addition{
	Meanwhile, the IBE still has challenges. 
	The results of the experiments also suggested the difficulty in dividing the state space to assign an explanation signal. 
	Prediction of environmental change in world with high complexity is still a challenging topic of research. 
}
Moreover, we fixed the spun of simulation $n=60$ in this paper, but in human communication, time granularity of the explanation differs depending on the context.
The problem of time granularity also occurs when an agent interprets the meaning of an instruction.
In future works, we wish to consider the appropriate time spun $n$ for the explanation of the behavior.
In addition, we are seeking the possibility of the application of IBE for agent's policy acquisition.
%
%
%
%
%
\balance{}

\balance{}

\bibliographystyle{acm-sigchi}
\bibliography{sample}

\end{document}